\documentclass[runningheads]{llncs}
\usepackage[T1]{fontenc}
\usepackage{graphicx}
\begin{document}
\title{A Comparison of Polynomial-Based Tree Clustering Methods}
%
%
\author{Pengyu Liu\inst{1} \and
Mariel V\'azquez\inst{2} \and
Nata\v{s}a Jonoska\inst{3}}
\authorrunning{P. Liu et al.}
%
\institute{University of Rhode Island, Kingston RI 02881, USA \and
University of California, Davis, Davis CA 95616, USA \and
University of South Florida, Tampa FL 33620, USA\\
\email{jonoska@usf.edu}}
\maketitle              
\begin{abstract}
Tree structures appear in many fields of the life sciences, including phylogenetics, developmental biology and nucleic acid structures.
Trees can be used to represent RNA secondary structures, which directly relate to the function of non-coding RNAs. 
Recent developments in sequencing technology and artificial intelligence have yielded numerous biological data that can be represented with tree structures. 
This requires novel methods for tree structure data analytics.
Tree polynomials provide a computationally efficient,  interpretable and comprehensive way to encode tree structures as matrices, which are compatible with most data analytics tools. Machine learning methods based on the Canberra distance between tree polynomials have been introduced to analyze phylogenies and nucleic acid structures. In this paper, we compare the performance of different distances in tree clustering methods based on a tree distinguishing polynomial. We also implement two basic autoencoder models for clustering trees using the polynomial. We find that the distance based methods with entry-level normalized distances have the highest clustering accuracy among the compared methods.

\end{abstract}

\section{Introduction}
Tree structures are important in many areas of life sciences. 
Phylogenetic trees record evolutionary information and patterns~\cite{Semple2003}. 
Tree representations capture the essence of RNA secondary structures~\cite{Schlick2018}.
Cell lineage trees record the developmental history of an organism from a stem cell~\cite{Santella2016}.
The advancements of sequencing technology and artificial intelligence applications in structure folding have produced a myriad of structural data~\cite{Kozlov2022,Abramson2024}. 
The abundance of structural data emphasizes the need for structural data analytics methods. 

In~\cite{Liu2021}, the author defined a complete polynomial invariant of trees.
This polynomial provides computationally efficient, interpretable and comprehensive methods for representing tree structures in a way compatible with data analytics tools.
We call this polynomial the {\em tree polynomial} or {\em polynomial encoding} of trees. 
Polynomial encoding together with distance-based machine learning methods have been introduced to infer evolutionary information~\cite{Liu2022} and to analyze RNA structures and predict the formation of non-canonical nucleic acid structures~\cite{Liu2024}.
In these applications, the Canberra distance~\cite{Lance1967} was implemented for measuring the similarity between trees. 
Here, we compare the performance of different distances in tree clustering with polynomial encoding. 
In addition, we implement two basic autoencoder models for clustering trees using tree polynomials and test their accuracy.

\section{Methods}

\subsection{Random Tree Generation}

We cluster random rooted binary trees generated by the beta-splitting model~\cite{Aldous1996}.
The model is parametrized with one parameter $\beta$.
We use three values $\beta=-1.5$, $\beta=-1$ and $\beta=0$ to generate random trees. 
These three values of $\beta$ correspond to the proportional to distinguishable arrangements (PDA) model, the Aldous branching model and the Yule model, respectively~\cite{Blum2006}, and they are known to generate trees with distinct topologies~\cite{Colijn2018}.
We choose these rooted binary trees to control the experiment. 
We generate 100 sets of random trees. 
Each set contains 300 random trees with 100 leaf vertices.
In each set, there are 100 trees generated by parameter $\beta=-1.5$, $\beta=-1$, and $\beta=0$, respectively, forming three generated clusters of trees.

\subsection{Tree Polynomial}

We use the tree distinguishing polynomial~\cite{Liu2021} to encode rooted trees. 
The definition of the polynomial is recursive from the leaf vertices to the root of a tree. 
Each leaf vertex $v$ has a polynomial $P(v,x,y) = x$.
An internal vertex $v$ with $k$ child vertices $v_1,v_2,\ldots,v_k$ has the polynomial defined by formula (\ref{eq1}).
\begin{equation}\label{eq1}
    P(v,x,y) = y + \prod_{i=1}^{k} P(v_i,x,y)
\end{equation}
The polynomial at the root is the polynomial that encodes the entire tree. 

\subsection{Polynomial-Based Tree Clustering Methods}

We can represent a tree polynomial $P(T,x,y)$ by its coefficient matrix $C_T$. 
The entry $c^{(i,j)}$ at the $(i-1)$'th row and $(j-1)$'th column of $C_T$ is the coefficient in the term $c^{(i,j)}x^iy^j$ of $P(T,x,y)$~\cite{Liu2022}.

\subsubsection{Tree Polynomial Distances with K-Medoids Clustering}

Here, we implement 6 distances of polynomial encoding and apply the k-medoids algorithm~\cite{Kaufman2005} for tree clustering. 
Let $P_1, P_2$ be two tree polynomials and $C_1, C_2$ be the corresponding coefficient matrices. 
We denote the entries in $C_1$ and $C_2$ by $c_1^{(i,j)}$ and $c_2^{(i,j)}$, respectively.

\paragraph{Euclidean distances}
The Euclidean distance and the normalized Euclidean distance between $C_1$ and $C_2$ are defined by formula (\ref{euc}) and (\ref{noreuc}), respectively.
\begin{equation}\label{euc}
    d_E(C_1,C_2) = \sqrt{\sum_{0 \leq i,j \leq n} (c_1^{(i,j)} - c_2^{(i,j)})^2}
\end{equation}
\begin{equation}\label{noreuc}
    d_{\bar{E}}(C_1,C_2) = \sqrt{\sum_{0 \leq i,j \leq n} \frac{(c_1^{(i,j)} - c_2^{(i,j)})^2}{\max(c_1^{(i,j)},c_2^{(i,j)})^2}}
\end{equation}

\paragraph{Manhattan distances}
The Manhattan distance and the normalized Manhattan distance between $C_1$ and $C_2$ are defined by formula (\ref{man}) and (\ref{norman}), respectively.
\begin{equation}\label{man}
    d_M(C_1,C_2) = \sum_{0 \leq i,j \leq n} |c_1^{(i,j)} - c_2^{(i,j)}|
\end{equation}
\begin{equation}\label{norman}
    d_{\bar{M}}(C_1,C_2) = \sum_{0 \leq i,j \leq n} \frac{|c_1^{(i,j)} - c_2^{(i,j)}|}{\max(c_1^{(i,j)},c_2^{(i,j)})}
\end{equation}

\paragraph{Canberra and Bray-Curtis distances}
The Canberra distance \cite{Lance1967} and the Bray-Curtis distance \cite{Bray1957} between $C_1$ and $C_2$ are defined by formula (\ref{can}) and (\ref{bc}), respectively.
\begin{equation}\label{can}
    d_C(C_1,C_2) = \sum_{0 \leq i,j \leq n} \frac{|c_1^{(i,j)} - c_2^{(i,j)}|}{(c_1^{(i,j)} + c_2^{(i,j)})}
\end{equation}
\begin{equation}\label{bc}
    d_{BC}(C_1,C_2) = \frac{\sum_{0 \leq i,j \leq n} |c_1^{(i,j)} - c_2^{(i,j)}|}{\sum_{0 \leq i,j \leq n} (c_1^{(i,j)} + c_2^{(i,j)})}
\end{equation}

The distances defined in (\ref{noreuc}), (\ref{norman}), (\ref{can}) are entry-level normalized. 
If the denominator in a summand equals $0$, then the summand is defined to be $0$.
For the distance defined in (\ref{bc}), the distance is $0$ if the denominator is $0$.

\paragraph{Clustering Experiment and Accuracy}
For each of the 100 sets of random trees, we compute the pairwise polynomial distance between the 100 random trees in the set using the 6 distances. 
This results in a 100-by-100 distance matrix for each of the distances. 
Then we apply the k-medoids algorithm on the distance matrix to perform tree clustering.
We compute the clustering accuracy by comparing the predicted clusters to the generated clusters using the majority rule.
We repeat the k-medoids clustering for 10 times to each distance matrix and compute the mean clustering accuracy. 

\subsubsection{Tree Polynomial Autoencoders with K-Means Clustering}
We construct two autoencoder neural network models~\cite{Baldi1989} to cluster tree polynomials.

\paragraph{Linear Autoencoder Model}
The encoder consists of 4 sequential linear layers with decreasing dimensions: 1024, 256, 64, and a latent layer of size 16.
Each of the first 3 layers is followed by a ReLU activation function. 
The decoder has a mirrored architecture, with a final sigmoid activation function. 
The model was trained using the Adam optimizer with learning rate 0.001 and mean squared error (MSE) loss over 500 epochs. 

\paragraph{Convolutional Autoencoder Model}
The encoder consists of 3 convolutional layers with 16, 32, and 64 filters, respectively, each with a $3\times3$ kernel, stride of 2, and padding of 1.
Each convolutional layer is followed by a ReLU activation function. 
The resulting feature maps are flattened and pass through a linear layer to produce a latent vector of size 16. 
The decoder mirrors this structure with a final sigmoid activation function.
The model was trained for 500 epochs using the Adam optimizer with a learning rate of 0.001 and MSE loss.

\paragraph{Clustering Experiment and Accuracy}
For each of the 100 sets of random trees, we use the autoencoder models to cluster tree polynomials of the 100 random trees in the set.
The coefficient matrix of each tree polynomial is normalized by dividing each entry with the maximum value of the matrix.
Note that for a tree with 100 leaf vertices, the coefficient matrix is of size $101\times 101$.
For the linear autoencoder, each normalized coefficient matrix is flattened to a vector of size $10201$.
For the convolutional autoencoder, the coefficient matrices stay unchanged.

For each autoencoder model, we extract the latent vector of size 16 for each of the random trees, and apply the k-means algorithm~\cite{Bishop2006} on the 100 latent vectors to perform tree clustering. 
We compute the clustering accuracy by comparing the predicted clusters to the generated clusters using the majority rule. 
We repeat the k-means clustering for 10 times to each set of 100 latent vectors and compute the mean clustering accuracy.

\section{Results}

\begin{figure}
\includegraphics[width=\textwidth]{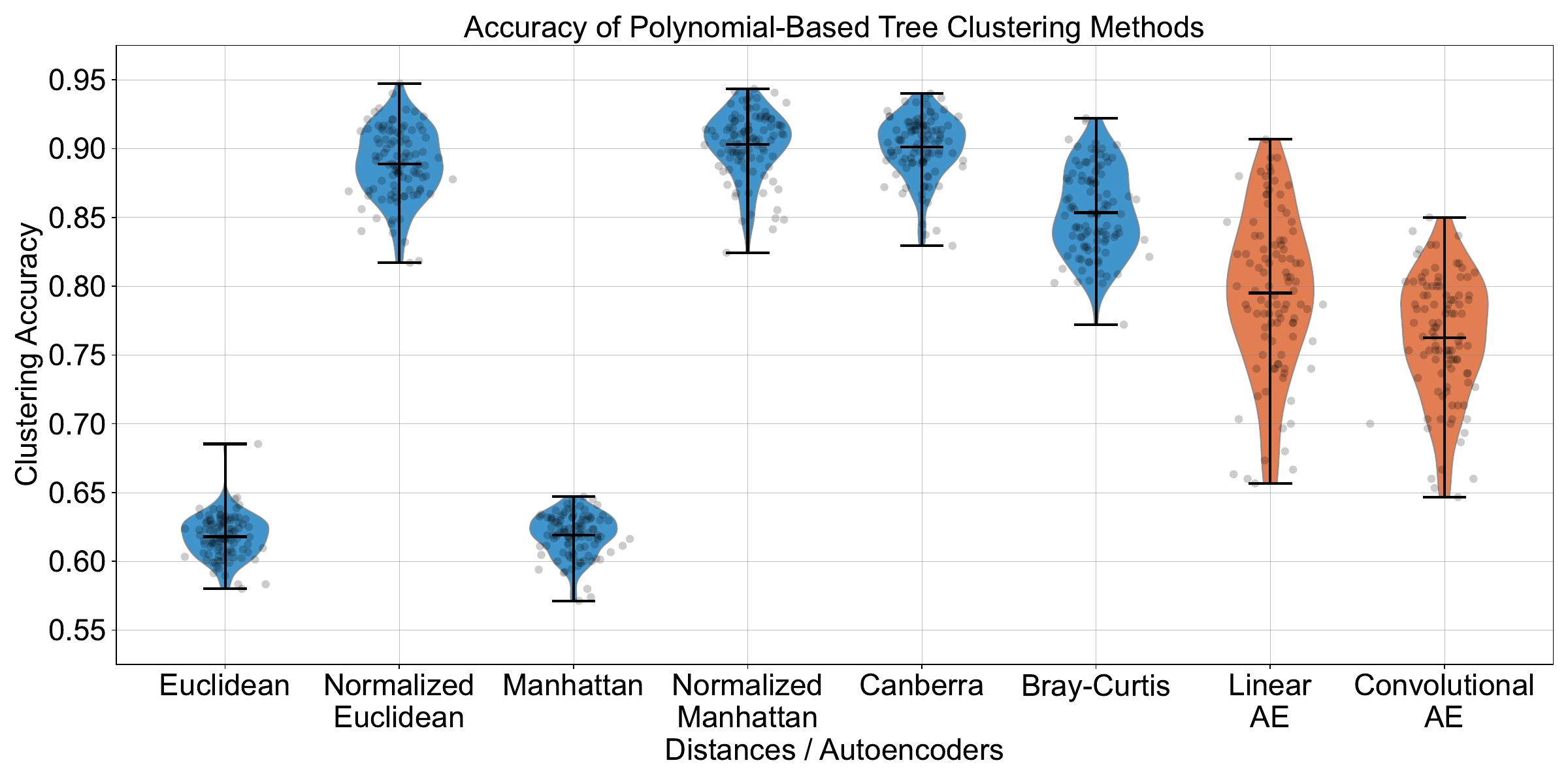}
\caption{This figure displays the distribution of the accuracy of 8 polynomial-based tree clustering methods. The distance-based methods are colored in blue, and the autoencoder-based methods are in red. Each dot in a violin plot represents the accuracy for one set of random trees. The bars in each violin plot shows the mean, maximum and minimum observed accuracy over the 100 sets of random trees.} \label{fig1}
\end{figure}

We display the clustering accuracy of the polynomial-based tree clustering methods in Fig.~\ref{fig1}.
The mean accuracy over the 100 sets of random trees is 0.62 for using the Euclidean distance in the polynomial-based tree clustering method, 0.89 for the normalized Euclidean distance, 0.62 for the Manhattan distance, 0.90 for the normalized Manhattan distance, 0.90 for the Canberra distance, and 0.85 for the Bray-Curtis distance. 
For clustering methods based on the linear and convolutional autoencoders, the mean accuracy is 0.79 and 0.76, respectively.

We observe that the distance-based methods with entry-level normalized distances have the highest accuracy among the compared tree clustering methods. 
These include the normalized Euclidean distance, the normalized Manhattan distance and the Canberra distance. 
The first two normalize the difference between each pair of corresponding entries in a coefficient matrix by the maximum of the pair. 
The Canberra distance normalizes the difference between each pair of corresponding entries by the sum of the pair. 
Methods based on the Euclidean and the Manhattan distance have low accuracy because the unnormalized distance between two coefficient matrices can be dominated by the large differences of a few entries. 

\section{Conclusion and future work}

We benchmarked the clustering accuracy of 8 polynomial-based tree clustering methods.
We found that distance-based methods with entry-level normalized distances have the highest accuracy. 

We used a small dataset of random rooted binary trees generated by the beta-splitting model with three different parameters, as they are known to have distinct tree topologies. 
Future work will include investigating stochastic processes that generate non-binary trees with distinct topologies and benchmarking the accuracy for clustering non-binary trees.
We will also test these methods for clustering large datasets of non-coding RNA secondary structures.

We introduced two basic autoencoder models for clustering trees using tree polynomials and tested their accuracy with the k-means clustering algorithm. 
These autoencoder-based methods did not perform as well as the distance-based methods with entry-level normalized distances.
This indicates a need for more structure data and novel neural network models more tailored for tree polynomials. 
The definition of tree polynomials depends on polynomial multiplication, so the coefficients of a tree polynomial are related. 
For example, the coefficients $a+b$ and $ab$ in $(x+ay)(x+by) = x^2 + (a+b)xy + aby^2$ are related.
This suggests that transformers or other attention-based models can be well-suited for performing structural data analytics using tree polynomials. 
We will explore these models and examine their performance in future work.

\begin{credits}
\subsubsection{\ackname} 
P.L. was supported by the startup funds of the University of Rhode Island. M.V. was supported by the NSF grant DMS/NIGMS\#2054347. N.J. was supported by NFS grant DMS/NIGMS\#2054321.

\subsubsection{\discintname}
The authors declare no competing interests. 
\end{credits}
%
%
%

\begin{thebibliography}{8}

\bibitem{Semple2003}
Semple, C., Steel, M.: Phylogenetics. Oxford University Press, New York (2003)

\bibitem{Schlick2018}
Schlick, T.: Adventures with RNA graphs. Methods \textbf{143}(1), 16--33 (2018)

\bibitem{Santella2016}
Santella, A., Kovacevic, I., Herndon, L.A., Hall, D.H., et al.: Digital development: a database of cell lineage differentiation in C. elegans with lineage phenotypes, cell-specific gene functions and a multiscale model. Nucleic Acids Research \textbf{44}(D1), D781--D785 (2016)

\bibitem{Kozlov2022}
Kozlov, A., Alves, J.M., Stamatakis, A., et al.: CellPhy: accurate and fast probabilistic inference of single-cell phylogenies from scDNA-seq data. Genome Biology \textbf{23}, 37 (2022)

\bibitem{Abramson2024}
Abramson, J., Adler, J., Dunger, J., et al.: Accurate structure prediction of biomolecular interactions with AlphaFold 3. Nature \textbf{630}, 493--500 (2024)

\bibitem{Liu2021}
Liu, P.: A tree distinguishing polynomial. Discrete Applied Mathematics \textbf{288}, 1--8 (2021)

\bibitem{Liu2022}
Liu, P., Biller, P., Gould, M., Colijn, C.: Analyzing phylogenetic trees with a tree lattice coordinate system and a graph polynomial. Systematic Biology \textbf{71}(6), 1378--1390 (2022)

\bibitem{Liu2024}
Liu, P., Lusk, J., Jonoska, N., Vázquez, M.: Tree polynomials identify a link between co-transcriptional R-loops and nascent RNA folding. PLOS Computational Biology \textbf{20}(12), e1012669 (2024)

\bibitem{Lance1967}
Lance, G.N., Williams, W.T.: A general theory of classificatory sorting strategies: II. Clustering systems. The Computer Journal \textbf{10}(3), 271--277 (1967)

\bibitem{Aldous1996}
Aldous, D.: Probability distributions on cladograms. In: Aldous, D., Pemantle, R. (eds) Random Discrete Structures, The IMA Volumes in Mathematics and its Applications, vol. 76, pp. 1--18. Springer, New York, NY (1996)

\bibitem{Blum2006}
Blum, M.G.B., François, O.: Which random processes describe the tree of life? A large-scale study of phylogenetic tree imbalance. Systematic Biology \textbf{55}(4), 685--691 (2006)

\bibitem{Colijn2018}
Colijn, C., Plazzotta, G.: A metric on phylogenetic tree shapes. Systematic Biology \textbf{67}(1), 113--126 (2018)

\bibitem{Kaufman2005}
Kaufman, L., Rousseeuw, P.J.: Finding groups in data: An introduction to cluster analysis. Wiley, Hoboken, NJ (2005)


\bibitem{Bray1957}
Bray, J.R., Curtis, J.T.: An ordination of the upland forest communities of southern Wisconsin. Ecological Monographs \textbf{27}(4), 325--349 (1957)

\bibitem{Baldi1989}
Baldi, P., Hornik, K.: Neural networks and principal component analysis: Learning from examples without local minima. Neural Networks \textbf{2}(1), 53--58 (1989)

\bibitem{Bishop2006}
Bishop, C.M.: Pattern recognition and machine learning. Springer, New York, NY (2006)



\end{thebibliography}
%

\end{document}